\documentclass[
]{ceurart}
\usepackage{listings}
\usepackage{caption}
\newcounter{promptcounter}
\newcommand{\promptcaption}[1]{%
    \refstepcounter{promptcounter}%
    \raggedright \textbf{Prompt \thepromptcounter:} #1
}
 
\begin{document}
%%
%% Rights management information.
%% CC-BY is default license.
\copyrightyear{2024}
\copyrightclause{Copyright for this paper by its authors.
  Use permitted under Creative Commons License Attribution 4.0
  International (CC BY 4.0).}
%%
%% This command is for the conference information
\conference{IRCDL'25: 21st conference on Information and Research science Connecting to Digital and Library science, Feb 20--21, 2025, Udine, Italy}
%%
%% The "title" command
\title{Bridging the Evaluation Gap: Leveraging Large Language Models for Topic Model Evaluation}
\tnotemark[1]
%\tnotetext[1]
\tnotetext[1]{
This work is jointly supported by the ``HybrInt - Hybrid Intelligence through Interpretable AI in Machine Perception and Interaction'' project (Zukunft Nds, Niedersächsisches Ministerium für Wissenschaft, Grant ID: ZN4219) and the \href{https://scinext-project.github.io/}{SCINEXT project} (BMBF, German Federal Ministry of Education and Research, Grant ID: 01lS22070).}
%%
%% The "author" command and its associated commands are used to define
%% the authors and their affiliations.
\author[1]{Zhiyin Tan}[%
orcid=0009-0002-4166-5810,
email=zhiyin.tan@l3s.de,
]
\cormark[1]
\address[1]{L3S Research Center, Leibniz University Hannover, Hannover, Germany}
\author[2]{Jennifer D'Souza}[%
orcid=0000-0002-6616-9509,
email=jennifer.dsouza@tib.eu,
]
\address[2]{TIB Leibniz Information Centre for Science and Technology, Hannover, Germany}
%% Footnotes
\cortext[1]{Corresponding author.}
%%
%% The abstract is a short summary of the work to be presented in the
%% article.
\begin{abstract}
This study presents a framework for automated evaluation of dynamically evolving topic taxonomies in scientific literature using Large Language Models (LLMs). In digital library systems, topic modeling plays a crucial role in efficiently organizing and retrieving scholarly content, guiding researchers through complex knowledge landscapes. As research domains proliferate and shift, traditional human-centric and static evaluation methods struggle to maintain relevance. The proposed approach harnesses LLMs to measure key quality dimensions—such as coherence, repetitiveness, diversity, and topic-document alignment—without heavy reliance on expert annotators or narrow statistical metrics. Tailored prompts guide LLM assessments, ensuring consistent and interpretable evaluations across various datasets and modeling techniques. Experiments on benchmark corpora demonstrate the method’s robustness, scalability, and adaptability, underscoring its value as a more holistic and dynamic alternative to conventional evaluation strategies.
\end{abstract}
\maketitle
\section{Introduction} %Highlights: interpretability, cross-domain datasets, model comparison, usefulness assessment
Topic taxonomies of Science have been traditionally used to simplify literature search, to study the structure and dynamics of scientific disciplines, or to facilitate bibliometric research evaluations~\cite{waltman2012new}. These taxonomies, often hierarchical and multidisciplinary, provide a framework for categorizing knowledge and can significantly influence the dissemination and evolution of scientific information \cite{moura2008proposal}. They play a crucial role in the organization of academic databases, directly impact the efficiency of information retrieval systems, and serve as essential tools for structuring and navigating vast repositories within digital library systems. As scientific output continues to grow exponentially, the need for effective and dynamically adaptable taxonomic systems \cite{dauxais2022towards,kotitsas2023scinobo} becomes increasingly important, not just for academic researchers but also for policymakers and funding agencies aiming to identify and support pivotal research areas \cite{bhat2014taxonomies}.

As we shift towards dynamically updatable taxonomies to manage the growing volumes of scientific literature, developing robust evaluation methods is essential to ensure their effectiveness \cite{dauxais2022towards}. These methods must verify that taxonomies adapt to the rapid evolution of scientific domains while consistently producing meaningful and coherent topics. Evaluating these dynamic systems involves assessing their accuracy in reflecting current research trends and their capacity to interlink related disciplines seamlessly \cite{shang2020nettaxo}. This process is crucial for maintaining the integrity and utility of taxonomies in facilitating efficient research discovery and supporting informed decision-making in scientific policy and funding strategies. Human evaluation is laborious and time-consuming; thus, we also need automated evaluation systems to efficiently manage and validate these complex, dynamic structures \cite{langlais2023rate}.

Building on the premise of automated evaluation systems, Large Language Models (LLMs) \cite{llm-amatriain-arxiv,llm-dsouza-orkg} are well-suited to function as evaluators of dynamic taxonomies due to their advanced natural language understanding and generation capabilities~\cite{stammbach2023revisiting}. Trained on extensive corpora, these models excel in discerning linguistic patterns and semantic relationships within complex datasets, making them ideal for assessing the coherence and relevance of topics generated by taxonomies. Moreover, LLMs offer a scalable, consistent, and context-sensitive approach to evaluation, overcoming key challenges of traditional methods such as reliance on human annotators or narrowly focused statistical metrics. With the ability to simulate nuanced human-like reasoning, LLMs can evaluate multiple dimensions of topic quality—such as coherence, repetitiveness, diversity, and topic-document alignment—while providing detailed, interpretable feedback.

In this work, we propose a novel framework that leverages LLMs as evaluators for topic model outputs, addressing key limitations in existing evaluation methodologies. The contributions of this work are threefold. First, we introduce a comprehensive set of metrics—coherence, repetitiveness, diversity, and topic-document alignment—that collectively capture multiple facets of topic model quality. Second, we design and implement tailored LLM prompts for each metric, ensuring consistency, interpretability, and adaptability across different datasets and topic modeling techniques. Third, we validate the framework through extensive experiments on benchmark datasets 20 Newsgroups (20NG)\footnote{http://qwone.com/~jason/20Newsgroups/} and a subset of scholarly documents from the International System for Agricultural
Science and Technology (AGRIS) \footnote{https://agris.fao.org/}, demonstrating its robustness and effectiveness.
This study not only provides a scalable and holistic solution for topic model evaluation but also paves the way for broader applications of LLMs in addressing dynamic and context-sensitive challenges in natural language processing. The code is available here\footnote{https://github.com/zhiyintan/topic-model-LLMjudgment}.
\begin{comment}
\begin{itemize}
\item Background on Topic Modeling (motivation, definition)
\item The importance and limitation of current evaluation methods
\item Propose to use LLM
\item The organization/contribution of the paper
\end{itemize}
\end{comment}
\section{Related Work}
\subsection{Topic Modeling Approaches and Their Evolution} % 1 paragraph: 5-6 sentences
The field of topic modeling has evolved significantly, advancing from early matrix factorization techniques to modern probabilistic and neural architectures. One foundational work, Latent Semantic Indexing (LSI) \cite{papadimitriou1998LSI},
utilized singular value decomposition to uncover conceptual associations between words and documents through co-occurrence patterns, though it lacked a fully generative probabilistic model.
Building on these concepts, Probabilistic Latent Semantic Indexing (pLSI) \cite{hofmann1999pLSI}
introduced a probabilistic framework where words are generated from a mixture of topics. This probabilistic perspective paved the way for LDA \cite{blei2003LDA}, 
a seminal contribution that introduced a fully generative Bayesian model capable of inferring a corpus-wide set of latent topics and their associated per-document distributions. LDA’s flexible yet tractable variational inference techniques (as well as alternative inference algorithms like Gibbs sampling, as demonstrated in \cite{Griffiths2004Gibbs})
solidified it as a cornerstone in topic modeling research.

In more recent years, researchers have embraced neural variational inference frameworks—inspired by VAE \cite{Kingma2014VAE}
—to develop neural topic models (NTMs) such as 
NVDM \cite{miao2016NVDM}, 
GSM/GSB/RSB
\cite{miao2017GSM-GSB-RSB}, 
ProdLDA/AVITM \cite{srivastava2017ProdLDA-AVITM}, 
and ETM \cite{dieng2020ETM}. 
These models employ continuous latent representations and deep neural networks to capture richer semantic structures and often rely on embeddings to represent words and documents. 
Another emerging direction leverages contextualized embeddings and large language models (LLMs) as building blocks or alternatives to traditional topic models. Methods such as clustering-based approaches \cite{sia2020tired-cluster},
CombinedTM \cite{bianchi2021CombinedTM}, and 
BERTopic \cite{grootendorst2022bertopic}
have demonstrated that directly clustering embeddings of documents and words can yield coherent, diverse topics. More recently, research efforts like
\cite{pham2024topicgpt, mu2024llmtm, mu2024llmtm-granularity-hallucination} propose employing LLMs directly to generate and refine topics, showing promise for overcoming certain limitations of classical topic modeling frameworks.
\subsection{Evaluation of Topic Models: Methods and Limitations}
Topic model evaluation has shifted from basic statistical measures to methods that capture human interpretability. Early work used held-out likelihood, introduced in \cite{blei2003LDA} 
and consistently employed by  
\cite{blei2007CTM},
measure how well a trained model can predict unseen data, although it can be computationally demanding. Similarly, log probability 
\cite{newman2009distributed} 
aggregates observed words and documents likelihood, providing an intuitive gauge of model fit but potentially favoring complexity over interpretability. Perplexity normalized held-out likelihood \cite{blei2003LDA, blei2007CTM, wang2008cDTM, newman2009distributed, Hinton2009UndirectedTM, srivastava2017ProdLDA-AVITM, ding2018coherenceNTM, card2018SCHOLAR, Zhang2018WHAI, dieng2020ETM}.
However, low perplexity does not necessarily translate to coherent or meaningful topics\cite{chang2009HumanTeaLeaves}, 
highlighting a fundamental disconnect between statistical quality and human interpretability.
To bridge this gap, human-centered evaluations like word intrusion and topic intrusion tasks \cite{chang2009HumanTeaLeaves}
were introduced, where human judges identify out-of-place words or topics. Human-rated coherence, first explored in
\cite{newman2010automaticHuman} 
and later adopted by 
\cite{mimno2011optimizing, aletras2013TC, stammbach2023revisitingLLM} 
asking annotators score topics on an ordinal scale. Although these methods closely reflect human understanding, their reliance on manual annotation constrains scalability and efficiency.
In response, automated metrics were then introduced to approximate human judgments. Topic words-based coherence metrics measure how strongly top-ranked topic words co-occur in the underlying data. Early approaches, such as coherence $C_{\text{UCI}}$ \cite{Newman2010TMforDL}
and coherence $C_{\text{UMass}}$ \cite{mimno2011optimizing},
rely on word co-occurrence frequencies and statistics derived from the training corpus. More refined metrics, like NPMI \cite{bouma2009NPMI}, normalize mutual information coherence scores $C_{\text{NPMI}}$ to better align with human judgments \cite{aletras2013TC, lau2014machineTeaLeaves, srivastava2017ProdLDA-AVITM, ding2018coherenceNTM, card2018SCHOLAR, wang2019ATM, wang2020BAT, grootendorst2022bertopic, wu2023ECRTM},
while coherence $C_{\text{v}}$ \cite{roder2015CV} 
uses a variation of $C_{\text{NPMI}}$ to compute topic coherence over a sliding window of size and adds a weight to assign more strength to more related words. Moreover, embedding-based coherence used by \cite{schnabel2015EmbeddingEvalTM, Nikolenko2016EmbeddingEvalTM, ding2018coherenceNTM, bianchi2021CombinedTM} further improves the match to human judgement \cite{Nikolenko2016EmbeddingEvalTM}.
Other metrics assess different aspects of topic quality beyond coherence. Diversity metrics ensure that the discovered topics are distinctive, not redundant or overlapping. For instance, topic diversity \cite{dieng2020ETM} 
counts the proportion of unique top words across topics, while topic redundancy \cite{Burkhardt2019TopicRedundancy} 
and topic uniqueness \cite{nan2019TopicUniqueness} 
measure how frequently top words appear across multiple topics. Similarly, inverted ranked-biased overlap \cite{bianchi2021CombinedTM} 
and embedding-based diversity metrics \cite{bianchi2021EmbedingCentroidTD, Terragni2021EmbedingTD} 
compare ranked word lists or semantic distances to ensure substantial topic variety. 
Document-level evaluations measure how well topics capture document's content. \cite{bhatia2017DocumentTopicEval} 
ask annotators to rate each topic’s relevance to a given document.
\cite{Korencic2018DocumenCoherence} vectorize documents associated with the selected topic and calculate a coherence score based on the document vectors.
\cite{bhatia2018topicIntrusionEval} 
tests whether an outlier topic can be identified given a document and a few topics.
Supervised coverage-based methods 
\cite{korenvcic2021CoverageEval}
match model-generated topics to a fixed, human-defined topics, though these methods are resource intensive.
More recently, LLM-based evaluations have emerged as a promising new paradigm. Studies \cite{stammbach2023revisitingLLM, rahimi2024contextualizedCoherence}
demonstrate that large language models can simulate human reasoning, providing nuanced judgments of topic coherence, word intrusions. 
\cite{yang2024llm} proposes a set of metrics to quantify the agreement between keywords generated from documents using LLM and topic words generated from documents by a topic model.
By leveraging LLMs’ extensive world knowledge and contextual reasoning, this approach overcomes the limitations of statistical co-occurrence, embedding similarities that often fail to capture semantic quality, and avoids the resource intensity of human-centric evaluations.
However, current LLM-based methods often focus on a single aspect, such as coherence. In contrast, our approach integrates multiple LLM-based metrics—including coherence, repetitiveness, diversity, and topic–document alignment—into a unified framework. Rather than simply measuring co-occurrence, our framework provides interpretable evidence (e.g., flagged outlier words and identified duplicate concepts) that explains topic flaws. Following \cite{chang2009HumanTeaLeaves, Blei2012PLDA}, topic model evaluations should assess the model's practical capabilities rather than rely on legacy metrics detached from its intended usage. Our novel topic–document alignment metrics explicitly reveal discrepancies between topic words and document content, which is crucial for applications like recommendation, summarization, and classification.
By integrating document-level and topic words-based assessments, our comprehensive, adversarially validated framework bridges the gap between statistical measures and human-centered evaluations, offering actionable insights to improve topic model performance.
\section{Our Solution: Evaluation Metrics and LLM as Evaluator\label{sec:llm-based evaluation metrics}}
In this section, we present our comprehensive evaluation framework that leverages LLMs as evaluators for topic models. We begin with topic words-based evaluation (Section~\ref{subsec:topic-words-evaluation}), where we introduce metrics to evaluate topic coherence $C_{\text{rate}}$ and $C_{\text{outlier}}$, and repetitiveness $R_{\text{rate}}$ and $R_{\text{duplicate}}$, as well include adversarial tests to validate metric robustness. Next, cross-topic evaluation (Section~\ref{subsec:cross-topic-evaluation}) focus on  topic diversity $D_{\text{rate}}$ that captures thematic distinctiveness across topics. Finally, topic–document alignment (Section~\ref{subsec:topic-document-alignment}) introduces novel metrics irrelevant topic words $A_{\text{ir-topic}}$ and missing themes $A_{\text{missing-theme}}$ to evaluate the correspondence between topic words and document content. Together, these subsections form an integrated, interpretable, and scalable framework for evaluating topic model performance.
\subsection{Topic words-based Evaluation\label{subsec:topic-words-evaluation}}
We evaluate individual topics by examining two key dimensions: (1) coherence, which measures the semantic consistency of the top-ranked topic words using the metrics \(C_{\text{rate}}\) and \(C_{\text{outlier}}\), and (2) repetitiveness, which assesses potential redundancy within the topic words using the metrics \(R_{\text{rate}}\) and \(R_{\text{duplicate}}\). This two-pronged evaluation enables us to quantify both the semantic integrity and the diversity of the generated topic words.
\paragraph{Coherence}
Topic coherence measures how well the top-ranked topic words form a semantically unified theme. Inspired by \cite{stammbach2023revisitingLLM},
we first employ an \textbf{coherence rating metric}, \(C_{\text{rate}}\), which asks the LLM to assess the overall semantic consistency of the topic words on a 3-point scale (with 1 indicating minimal alignment and 3 indicating strong coherence). While \(C_{\text{rate}}\) yields an overall numerical score, it does not reveal which and how many specific words are responsible for any lack of coherence. To enhance interpretability, we introduce an auxiliary \textbf{outlier detection metric}, $C_{\text{outlier}}$, that explicitly identifies semantic outlier words. In this procedure, the LLM extracts candidate outliers over 5 iterations, and a word flagged in at least 3 out of 5 iterations is deemed a semantic outlier. We then count the number of outliers as the final evaluation result, and the outlier words themselves are saved for later case studies. In addition, we perform an \textbf{adversarial test} $AdvT_{\text{outlier}}$ to validate the reliability of the outlier detection inspired by established word intrusion methodologies \cite{stammbach2023revisitingLLM, chang2009HumanTeaLeaves}.
A semantically unrelated term (e.g., \textit{"Shakespeare"}) is inserted into the topic list, and the LLM is expected to correctly identify the inserted term. Even if the LLM flags additional words along with the inserted term, the detection is considered successful. Each successful detection is assigned a score of 1 and each unsuccessful attempt a score of 0. The final result is calculated as the percentage of successful detections over the total number of tests. Prompt~\ref{tab:prompt_coherence} presents the prompts for the two evaluation metrics as well as for the adversarial test.
\begin{table}[]
\promptcaption{Prompt for evaluating coherence.}
\label{tab:prompt_coherence}
\begin{tabular}{p{15cm}}
\textbf{Prompt for coherence rating metric} $C_{\text{rate}}$\\
Given a list of words {[}TOPIC WORDS{]} representing a topic, assess the degree of semantic consistency among the words in the context of the topic. Consider the remaining words in the list as the contextual basis for each word's semantics.
Rate the coherence of the topic on a scale of 1 to 3, where 1 indicates that the words are mostly unrelated, and 3 indicates that the words are highly related and form a clear, unified theme.\\
The rate is: [RATE]\\  \hline
\textbf{Prompt for outlier detection metric} $C_{\text{outlier}}$\\
Given a list of words {[}TOPIC WORDS{]} representing a topic. Consider the remaining words in the list as the contextual basis for each word's semantics. Which words are not semantically consistent with the remaining words and put them into a comma-separated list. \\
The semantically inconsistent words are: [WORD LIST]\\  \hline
\textbf{Prompt for adversarial test of outlier detection metric} $AdvT_{\text{outlier}}$\\
Which words from this list {[}TOPIC WORDS{]} are not semantically consistent with the remaining words? \\
The semantically inconsistent words are: [WORD LIST]\\
\end{tabular}
\end{table}
\paragraph{Repetitiveness}
While coherence focuses on thematic alignment, we introduce a \textbf{repetitiveness rating metric} $R_{\text{rate}}$ to assess whether the perceived coherence is due to redundant topic words. $R_{\text{rate}}$ is rating on a 3-point scale, where a rating of 1 indicates high repetitiveness with significant semantic overlap, and a rating of 3 indicates minimal repetition with diverse and distinctive words. To further elucidate these ratings, we introduce an auxiliary \textbf{duplicate concept detection metric}, \(R_{\text{duplicate}}\), which explicitly identifies exact semantic repetitions in the topic word list. \(R_{\text{duplicate}}\) is critical as it helps distinguish genuine topic coherence from inflated scores due to redundancy. For each topic word, we compute a binary indicator: if a word has at least one other conceptual repetition in the list, it is assigned a value of 1, otherwise, 0. The final \(R_{\text{duplicate}}\) score for a topic is the sum of these indicators, reflecting the number of topic words that have at least one duplicate in the list. 
In addition, we perform an \textbf{adversarial test} $AdvT_{\text{duplicate}}$ to validate the reliability of using the LLM for duplicate concept detection. In this test, we randomly select an anchor word from the topic word list and manually choose a conceptually identical word to serve as its duplicate. We then insert this duplicate into the topic word list. The LLM is expected to identify the inserted duplicate given the anchor word. Each successful detection is scored as 1, while an unsuccessful one is scored as 0.
Prompt~\ref{tab:prompt_repetitive} presents the prompts designed for both \(R_{\text{rate}}\) and \(R_{\text{duplicate}}\), as well as for the associated adversarial test.
\begin{table}[]
\promptcaption{Prompt for evaluating repetitiveness.}
\label{tab:prompt_repetitive}
\begin{tabular}{p{15cm}}
\textbf{Prompt for repetitiveness rating metric} $R_{\text{rate}}$\\
Given a list of words {[}TOPIC WORDS{]} representing a topic, evaluate if there are words that are semantically equivalent.
Rate the repetitive on a scale of 1 to 3, where 1 indicates highly repetitive with significant semantic overlap, and 3 indicates minimal repetition with diverse and distinctive words. \\
The rate is: [RATE]\\  \hline
\textbf{Prompt for duplicate concept detection metric} $R_{\text{duplicate}}$\\
Given a list of words {[}TOPIC WORDS{]} representing a topic, identify pairs of words that refer to the exact same concept or idea (not just related or similar). Provide each pair as a tuple in a comma-separated list. \\
The word pairs are: [WORD LIST] \\ \hline
\textbf{Prompt for adversarial test of duplicate concept detection metric} $AdvT_{\text{duplicate}}$\\
Given a list of words {[}TOPIC WORDS{]} representing a topic, which words from the list have the exact same concept or idea (not just related or similar)? \\
The word pair with [ANCHOR] is ('None' or a word): [None/WORD]
\end{tabular}
\end{table}
\subsection{Cross-topic Evaluation\label{subsec:cross-topic-evaluation}}
\paragraph{Diversity}
Diversity  quantifies the uniqueness among generated topics by assessing the thematic distinctiveness of their associated top words. Inspired by \cite{Terragni2021EmbedingTD}'s word embedding-based pairwise distance, we exhaustively extract all possible pairs of topics, with each topic represented by its corresponding topic word list. For each pair, the LLM rates the thematic distinctiveness on a 3-point scale, where a rating of 1 denotes partial overlap (low diversity) and a rating of 3 denotes minimal overlap (high distinctiveness). Finally, the average of all pairwise ratings is computed to yield the overall \textbf{diversity rating}, \(D_{\text{rate}}\).
The prompt for diversity evaluation is provided in Prompt~\ref{tab:prompt_diversity}.
\begin{table}[]
\promptcaption{Prompt for evaluating diversity.}
\label{tab:prompt_diversity}
\begin{tabular}{p{15cm}}
\textbf{Prompt for diversity rating metric} $D_{\text{rate}}$\\
Given two groups of topic words: {[}TOPIC WORDS 1{]}, {[}TOPIC WORDS 2{]}, analyze the themes represented by the two groups.
Rate on a scale of 1-3 based on the degree of thematic distinctiveness between the two groups:
Rate 1: Partial overlapping themes.
Rate 3: Highly distinctive themes.\\
The rate is: [RATE]
\end{tabular}
\end{table}
\subsection{Topic-document Alignment\label{subsec:topic-document-alignment}}
Document-level evaluation focuses on assessing how effectively topics capture the underlying themes of documents. Early methods relied on human annotations~\cite{bhatia2017DocumentTopicEval, bhatia2018topicIntrusionEval} or supervised matching against curated references~\cite{korenvcic2021CoverageEval} to measure topic relevance. However, these approaches are resource-intensive and lack scalability. Recent work by \cite{yang2024llm} introduces a set of metrics that quantify the agreement between keywords generated by LLMs from documents and the topic words produced by topic models. Although these metrics capture similarity, they do not quantify the degree of mismatch between the two sets. This unaccounted discrepancy is critical for evaluating how well a topic model covers less frequent or nuanced themes, which are often key to understanding long-tail phenomena. By incorporating measures of mismatch, we can gain a more complete picture of the model’s limitations and identify specific areas where the topic representation may require further improvement. Motivated by these limitations, we propose two novel LLM-based metrics for topic-document alignment: irrelevant topic words detection metric $A_{\text{ir-topic}}$ and missing themes detection metric $A_{\text{missing-theme}}$. These metrics leverage the contextual understanding of LLMs to assess both overrepresentation (irrelevant topic words) and underrepresentation (missing themes) in topic-document relationships, providing a more comprehensive evaluation of how well topics capture document content.
Prompts for these evaluations are provided in Prompt~\ref{tab:prompt_topic_document_alignment}.
\paragraph{Irrelevant Topic Words Detection} Metric $A_{\text{ir-topic}}$ assesses the extent to which a topic contains words that are not relevant to the content of its associated documents. For each topic-document pair, we instruct the LLM to identify topic words that are not explicitly or implicitly related to the document. The number of extraneous words is tallied for each document and then averaged across all pairs, providing a precise measure of overrepresentation. 
\paragraph{Missing Themes Detection} Conversely, metric $A_{\text{missing-theme}}$ quantifies the extent to which a topic fails to capture key themes present in the documents. For each topic-document pair, the LLM extracts significant themes from the document that are absent in the topic word list and counts these missing themes. The resulting counts are then averaged across all pairs, yielding a measure of underrepresentation. \\
\begin{table}[]
\promptcaption{Prompt for evaluating topic-document alignment.}
\label{tab:prompt_topic_document_alignment}
\begin{tabular}{p{15cm}}
\textbf{Prompt for irrelevant topic words detection metric} $A_{\text{ir-topic}}$\\
Given a document: {[}DOCUMENT{]} and a topic word list {[}TOPIC WORDS{]}, identify which topics in the word list are not relevant to the document. \\
Return these extraneous topics, or [ ] if all topics in the word list are relevant to the document. \\
Return the extraneous topics list or [ ]: [TOPIC WORDS/[ ]]\\ \hline
\textbf{Prompt for missing themes detection metric} $A_{\text{missing-theme}}$\\
Given a document: {[}DOCUMENT{]} and a topic word list {[}TOPIC WORDS{]}, identify which themes present in the document are not included in the topic word list. \\
Return these missing themes, or [ ] if all themes from the document are included in the word list. \\
Return the missed themes list or [ ]: [MISSING THEMES/[ ]]\\
\end{tabular}
\end{table}
\section{Experimental Setup}
\subsection{Data}
\paragraph{Datasets} 
The \textbf{20NG} dataset is a widely used benchmark comprising approximately 20,000 newsgroup posts organized into 20 categories. We adopt the pre-processed version from OCTIS\footnote{https://github.com/MIND-Lab/OCTIS/}, which contains 11,415 training and 4,894 test documents. Known for its diverse topics, 20NG has been extensively employed in topic model evaluations~\cite{Hinton2009UndirectedTM, miao2017GSM-GSB-RSB, Zhang2018WHAI, wu2023ECRTM}. 
Besides, from a repository of approximately 14 million records of food and agricultural scholarly documents—we extract a subset by excluding non-English documents, those with titles shorter than five tokens or abstracts shorter than 40 tokens, and duplicate records (by DOI, and named it \textbf{AGRIS}.
The final dataset comprises 50,067 documents (45,060 for training and 5,007 for testing). % 72,286 
For each document, we retain the title and abstract. %, and AGROVOC keywords. 
To support sentence-level analysis, abstracts are segmented using SaT~\cite{frohmann-etal-2024-segment}, yielding 454,850 training and 50,703 test entries.
This granularity allows multiple topics to be assigned to a single document, reflecting the multi-faceted nature of scholarly texts, where a single work often spans diverse thematic areas.
\paragraph{Domain-Specific Stopword Removal}
Stopword removal is a critical preprocessing step, particularly for domain-specific data. Stopwords are frequent but low-information terms, and their removal, guided by Zipf’s law \cite{george1949zipf}, reduces token counts while preserving vocabulary diversity, optimizing computational efficiency without compromising semantic integrity.
Generic stopword lists often overlook contextually irrelevant terms in specialized domains. For AGRIS, we employed an information-theoretic framework \cite{Gerlach2019} to identify and remove domain-specific stopwords. The 20NG dataset, pre-processed by OCTIS, required no further stopword refinement.
\subsection{Topic Models}
We evaluated four topic models chosen for their methodological diversity and proven performance, spanning traditional probabilistic approaches to neural and embedding-based methods, enabling a comprehensive comparison. Their key characteristics and implementations are detailed below.
\begin{itemize}
\item \textit{\textbf{Latent Dirichlet Allocation (LDA)}} \cite{blei2003LDA}: a foundational probabilistic topic model that represents documents as mixtures of topics, with each topic modeled as a distribution over words. We use the Gensim implementation\footnote{https://radimrehurek.com/gensim/models/ldamodel.html}.
\item \textit{\textbf{Product of Experts Latent Dirichlet Allocation (ProdLDA)}} \cite{srivastava2017ProdLDA-AVITM}: a neural adaptation of LDA that leverages variational autoencoders to enhance scalability and improve topic coherence. we adopt the code provided by the TopMost toolkit\footnote{https://github.com/BobXWu/TopMost}.
\item \textit{\textbf{CombinedTM}} \cite{bianchi2021CombinedTM}: it integrates contextual embeddings from pre-trained transformers into the LDA framework, effectively capturing semantic nuances through deep neural embeddings. We used the official implementation\footnote{https://github.com/MilaNLProc/contextualized-topic-models}.
\item \textit{\textbf{BERTopic}} \cite{grootendorst2022bertopic}: combines document embeddings with a class-based TF-IDF procedure to generate coherent and interpretable topics. For this study, we configured BERTopic with UMAP for dimensionality reduction and HDBSCAN for clustering, following its standard pipeline\footnote{https://maartengr.github.io/BERTopic/}.
\end{itemize}
For each model, we conducted an extensive parameter tuning process to identify the optimal settings for two key evaluation metrics: $C_{\text{v}}$ \cite{roder2015CV} for topic coherence and $D_{\text{unique}}$ \cite{dieng2020ETM} for topic diversity. 
Once the optimal configurations were determined, we obtain the top 10 topic words and the topic-document pairs from each model on both the 20NG and AGRIS datasets with number of topic $K=50$ and $K=100$. Each configuration was run ten times to account for variability in probabilistic and neural-based outputs. 
This resulted in ten aggregated sets of results for each model and dataset, ensuring a robust and statistically sound evaluation. This rigorous evaluation protocol not only ensures a fair comparison across the diverse modeling paradigms but also provides comprehensive insights into each model’s strengths and limitations in capturing thematic content across varied datasets.
\subsection{Evaluation}
\paragraph{Metrics} We employ two widely recognized automated metrics as baselines: the coherence metric $C_{\text{v}}$ \cite{roder2015CV}, which measures the semantic consistency of top-ranked topic words,
and the diversity metric $D_{\text{unique}}$ \cite{dieng2020ETM}, which quantifies the proportion of unique words across all topics.
To address the limitations of automated metrics, we also use the suite of LLM-based metrics described in Section~\ref{sec:llm-based evaluation metrics} for a more nuanced evaluation of topic quality. These include:
(1) \textbf{Coherence Metrics}: coherence rating metric $C_{\text{rate}}$ and outlier detection metric $C_{\text{outlier}}$.
(2) \textbf{Repetitiveness Metrics}: repetitiveness rating metric $R_{\text{rate}}$ and duplicate concept detection metric $R_{\text{duplicate}}$.
(3) \textbf{Diversity Metric}: diversity
rating metric $D_{\text{rate}}$.
(4) \textbf{Topic-document Alignment Metrics}: irrelevant topic words detection metric $A_{\text{ir-topic}}$ and missing themes detection metric $A_{\text{missing-theme}}$.
For the topic-document alignment metrics, a sample set was constructed by randomly selecting one iteration of the model's output and sampling up to 100 associated documents per topic—yielding 59,499 samples from AGRIS and 38,321 from 20NG—thus ensuring a comprehensive evaluation.
These metrics harness LLMs' deep semantic understanding to provide a comprehensive, multi-dimensional evaluation of topic quality.
\paragraph{LLM as Evaluators}
We selected three open-source LLMs as evaluators for the proposed metrics:
\textit{Mistral-7B-Instruct-v0.3}\footnote{https://huggingface.co/mistralai/Mistral-7B-Instruct-v0.3} (referred to as Mistral), 
\textit{Meta-Llama-3.1-8B-Instruct}\footnote{https://huggingface.co/meta-llama/Llama-3.1-8B-Instruct} (referred to as Llama), and
\textit{Qwen2.5-14B-Instruct}\footnote{https://huggingface.co/Qwen/Qwen2.5-14B-Instruct} (referred to as Qwen).
These LLMs, chosen for their diverse pretraining corpora and instruction-tuning objectives, exhibit robust semantic understanding. Their complementary strengths ensure reliable, scalable, and reproducible evaluations, while promoting transparency and facilitating replication in the research community.
\paragraph{Efficiency and Scalability}
We evaluated the computational efficiency and scalability of our LLM-based evaluation framework. All experiments were conducted on an NVIDIA A100 GPU, with approximately 40\,GB allocated for coherence, repetitiveness, and diversity evaluations, and about 70\,GB for topic-document alignment. The combined evaluation (three LLMs) time for coherence, repetitiveness, and diversity ranged from 15 minutes for $K=50$ to 35 minutes for $K=100$, while topic-document alignment evaluation required between 2 and 3 hours. On GPUs with lower performance and memory, reducing the batch size enables smooth operation at the expense of increased processing time, while parallel computing can reduce runtime proportionally to the number of GPUs available. These results demonstrate that our framework is both computationally efficient and scalable, making it suitable for extensive evaluations of topic models.
\section{Results and Discussion}
\subsection{Quantitative Results}
\paragraph{Adversarial Test}
We sampled 100 topics (each with 10 words) from four topic models applied to the 20NG and AGRIS datasets. For adversarial test of outlier detection $AdvT_{\text{outlier}}$, the success rates on 20NG are 
77\% (Mistral), 81\% (Llama), and 90\% (Qwen), 
and on AGRIS, 82\% (Mistral), 85\% (Llama), and 93\% (Qwen).
For duplicate concept detection $AdvT_{\text{duplicate}}$, success rates on 20NG are, 
37\% (Mistral), 81\% (Llama), and 84\% (Qwen), and on AGRIS, 29\% (Mistral), 74\% (Llama), and 81\% (Qwen).
These results highlight significant variability among LLM evaluators and underscore the importance of using multiple evaluators to reliably assess topic quality.
\paragraph{Coherence} 
Tables~\ref{tab:coherence_20ng} and~\ref{tab:coherence_agris} indicate that, based on the coherence rating metric \(C_{\text{rate}}\), \textit{BERTopic} consistently outperforms the other topic models on both datasets, which aligns with the automated metric \(C_{\text{v}}\). With respect to the outlier detection metric \(C_{\text{outlier}}\), for \(K=50\) two of the three LLMs report the fewest outliers for \textit{BERTopic}, supporting its superior coherence. At \(K=100\), Qwen maintain its preference for \textit{BERTopic} while Mistral finds that \textit{LDA} and \textit{CombinedTM} are comparable. In contrast, Llama’s \(C_{\text{outlier}}\) suggest that \textit{BERTopic} has more outliers. These discrepancies are further analysed in Section~\ref{subsec:qualitative_analysis}.
\begin{table}[]
    \caption{Evaluation results of coherence for 20NG: automated metric $C_{\text{v}}$ vs. LLM-based metrics $C_{\text{rate}}$ and $C_{\text{outlier}}$.}
    \label{tab:coherence_20ng}
\centering
\resizebox{\textwidth}{!}{%
\begin{tabular}{llcccccc|lcccccc}
\hline
Coherence & & \multicolumn{6}{c|}{$K=50$} & & \multicolumn{6}{c}{$K=100$} \\ \cline{2-15} 
          & \multicolumn{1}{c|}{$C_{\text{v}}$} & \multicolumn{3}{c|}{$C_{\text{rate}}$  ↑} & \multicolumn{3}{c|}{$C_{\text{outlier}}$ ↓} 
          & \multicolumn{1}{c|}{$C_{\text{v}}$} & \multicolumn{3}{c|}{$C_{\text{rate}}$  ↑} & \multicolumn{3}{c}{$C_{\text{outlier}}$ ↓}  \\ \cline{3-8} \cline{10-15} 
          & \multicolumn{1}{c|}{} & Mistral & Llama& \multicolumn{1}{c|}{Qwen} & Mistral & Llama& Qwen 
          & \multicolumn{1}{c|}{} & Mistral & Llama  & \multicolumn{1}{c|}{Qwen} & Mistral & Llama& Qwen \\ \hline
\textit{LDA} & \multicolumn{1}{c|}{0.558} & 2.140 & 2.390 & \multicolumn{1}{c|}{2.047} & 0.678 & \textbf{1.274} & 2.104
    & \multicolumn{1}{c|}{0.508} & 2.111 & 2.282 & \multicolumn{1}{c|}{1.881} & \textbf{0.700} & 1.344 & 2.334 \\
\textit{ProdLDA} & \multicolumn{1}{c|}{0.514} & 2.078 & 2.204 & \multicolumn{1}{c|}{1.730} & 1.152 & 1.308 & 2.318 
        & \multicolumn{1}{c|}{0.470} & 2.053 & 2.132 & \multicolumn{1}{c|}{1.550} & 1.205 & \textbf{1.239} & 2.376 \\
\textit{CombinedTM}  & \multicolumn{1}{c|}{0.560} & 2.088 & 2.366 & \multicolumn{1}{c|}{2.086} & 0.778 & 1.376 & 2.108
            & \multicolumn{1}{c|}{\textbf{0.538}} & 2.060 & 2.329 & \multicolumn{1}{c|}{1.955} & 0.754 & 1.399 & 2.228 \\
\textit{BERTopic}    & \multicolumn{1}{c|}{\textbf{0.585}} & \textbf{2.277} & \textbf{2.487} & \multicolumn{1}{c|} {\textbf{2.421}} & \textbf{0.602} & 1.471 & \textbf{1.915} 
            & \multicolumn{1}{c|}{0.534} & \textbf{2.159} & \textbf{2.416} & \multicolumn{1}{c|}{\textbf{2.160}} & 0.842 & 1.423 & \textbf{2.137} \\ \hline
\end{tabular}%
}
\end{table}
\begin{table}[]
    \caption{Evaluation results of coherence for AGRIS: automated metric $C_{\text{v}}$ vs. LLM-based metrics $C_{\text{rate}}$ and $C_{\text{outlier}}$.}
    \label{tab:coherence_agris}
\centering
\resizebox{\textwidth}{!}{%
\begin{tabular}{llcccccc|lcccccc}
\hline
Coherence & & \multicolumn{6}{c|}{$K=50$} & & \multicolumn{6}{c}{$K=100$} \\ \cline{2-15} 
         & \multicolumn{1}{c|}{$C_{\text{v}}$} & \multicolumn{3}{c|}{$C_{\text{rate}}$  ↑} & \multicolumn{3}{c|}{$C_{\text{outlier}}$ ↓} 
          & \multicolumn{1}{c|}{$C_{\text{v}}$} & \multicolumn{3}{c|}{$C_{\text{rate}}$  ↑} & \multicolumn{3}{c}{$C_{\text{outlier}}$ ↓}  \\ \cline{3-8} \cline{10-15} 
          & \multicolumn{1}{c|}{} & Mistral & Llama& \multicolumn{1}{c|}{Qwen} & Mistral & Llama& Qwen 
          & \multicolumn{1}{c|}{} & Mistral & Llama  & \multicolumn{1}{c|}{Qwen} & Mistral & Llama& Qwen \\ \hline
          
\textit{LDA} & \multicolumn{1}{c|}{0.411} & 2.106 & 2.390 & \multicolumn{1}{c|}{2.113} & 0.778 & 1.32  & 2.436 
    & \multicolumn{1}{c|}{0.353} & 2.029 & 2.309 & \multicolumn{1}{c|}{1.894} & 0.961 & \textbf{1.336} & 2.462 \\
\textit{ProdLDA} & \multicolumn{1}{c|}{0.557} & 2.214 & 2.356 & \multicolumn{1}{c|}{2.442} & 1.162 & 1.428 & 2.088 
        & \multicolumn{1}{c|}{0.552} & 2.148 & 2.356 & \multicolumn{1}{c|}{2.38}  & 1.207 & 1.407 & 2.265 \\
\textit{CombinedTM}  & \multicolumn{1}{c|}{0.507} & 2.172 & 2.442 & \multicolumn{1}{c|}{2.424} & 0.924 & \textbf{1.318} & 2.234 
            & \multicolumn{1}{c|}{0.571} & 2.186 & 2.486 & \multicolumn{1}{c|}{2.457} & \textbf{0.715} & 1.366 & 2.173 \\
\textit{BERTopic}    & \multicolumn{1}{c|}{\textbf{0.655}} & \textbf{2.544} & \textbf{2.672} & \multicolumn{1}{c|} {\textbf{2.878}} & \textbf{0.679} & 1.558 & \textbf{1.665} 
            & \multicolumn{1}{c|}{\textbf{0.627}} & \textbf{2.452} & \textbf{2.634} & \multicolumn{1}{c|} {\textbf{2.777}} & 0.805 & 1.477 & \textbf{1.684} \\ \hline
\end{tabular}%
}
\end{table}
\paragraph{Repetitiveness} 
To discern whether high coherence reflects genuine semantic unity or is driven by redundant topic words, we examine $R_{\text{rate}}$ and $R_{\text{duplicate}}$. A robust coherence result should be accompanied by a high $R_{\text{rate}}$ (indicating minimal repetition) and a low $R_{\text{duplicate}}$ (indicating few duplicate concepts). 
For instance, although \textit{BERTopic} shows high coherence, Tables~\ref{tab:repetitive_20ng} and \ref{tab:repetitive_agris} reveal that, most LLM evaluators report a low $R_{\text{rate}}$ and a high $R_{\text{duplicate}}$ for it, suggesting its apparent coherence may be inflated by redundant word selection.
In contrast, \textit{ProdLDA} on 20NG and \textit{LDA} on AGRIS tend to exhibit relatively high $R_{\text{rate}}$ and low $R_{\text{duplicate}}$ scores, implying that their coherence is less likely to be enhanced by redundant word selection. 
These findings underscore the importance of assessing repetitiveness alongside coherence to ensure that high coherence truly reflect meaningful topic quality.
\begin{table}[]
    \caption{Evaluation results of repetitiveness for 20NG: LLM-based metrics $R_{\text{rate}}$ and $R_{\text{duplicate}}$.}
    \label{tab:repetitive_20ng}
\centering
\resizebox{\textwidth}{!}{%
\begin{tabular}{lcccccc|cccccc}
\hline
Repetitiveness & \multicolumn{6}{c|}{$K=50$} & \multicolumn{6}{c}{$K=100$} \\ \cline{2-13} 
        & \multicolumn{3}{c|}{$R_{\text{rate}}$  ↑} & \multicolumn{3}{c|}{$R_{\text{duplicate}}$ ↓} 
        & \multicolumn{3}{c|}{$R_{\text{rate}}$  ↑} & \multicolumn{3}{c}{$R_{\text{duplicate}}$ ↓} \\ \cline{2-13} 
    & Mistral & Llama& \multicolumn{1}{c|}{Qwen} & Mistral & Llama& Qwen 
    & Mistral & Llama& \multicolumn{1}{c|}{Qwen} & Mistral & Llama& Qwen  \\ \hline
\textit{LDA} & 2.000 & 1.652 & \multicolumn{1}{c|}{2.091} & 2.333 & 3.586 & \textbf{1.147} 
    & 2.000   & 1.641 & \multicolumn{1}{c|}{2.120} 
    & 2.208 & 3.673 & \textbf{1.139} \\
\textit{ProdLDA} & 2.000 & \textbf{1.698} & \multicolumn{1}{c|}{\textbf{2.204}} 
        & \textbf{2.112} & \textbf{3.566} & 1.682 & 2.000   
        & \textbf{1.723} & \multicolumn{1}{c|}{\textbf{2.248}} 
        & \textbf{2.022} & 3.597 & 1.585 \\
\textit{CombinedTM}  & 2.000 & 1.650 & \multicolumn{1}{c|}{2.132} & 2.348 
            & 3.578 & 1.442 & 2.000   & 1.670 & \multicolumn{1}{c|}{2.131} 
            & 2.259 & \textbf{3.518} & 1.291 \\
\textit{BERTopic}    & 2.000 & 1.595 & \multicolumn{1}{c|}{2.034} & 2.598 
            & 3.686 & 1.379 & 2.000   & 1.639 & \multicolumn{1}{c|}{2.064} 
            & 2.459 & 3.575 & 1.377 \\ \hline
\end{tabular}%
}
\end{table}
\begin{table}[]
    \caption{Evaluation results of repetitiveness for AGRIS: LLM-based metrics $R_{\text{rate}}$ and $R_{\text{duplicate}}$.}
    \label{tab:repetitive_agris}
\centering
\resizebox{\textwidth}{!}{%
\begin{tabular}{lcccccc|cccccc}
\hline
Repetitiveness & \multicolumn{6}{c|}{$K=50$} & \multicolumn{6}{c}{$K=100$} \\ \cline{2-13} 
           & \multicolumn{3}{c|}{$R_{\text{rate}}$  ↑} & \multicolumn{3}{c|}{$R_{\text{duplicate}}$ ↓}
           & \multicolumn{3}{c|}{$R_{\text{rate}}$  ↑} & \multicolumn{3}{c}{$R_{\text{duplicate}}$ ↓} \\ \cline{2-13} 
           & Mistral & Llama& \multicolumn{1}{c|}{Qwen}  & Mistral        & Llama& Qwen  & Mistral        & Llama& \multicolumn{1}{c|}{Qwen}  & Mistral        & Llama& Qwen  \\ \hline
\textit{LDA} & 2.000 & \textbf{1.710} & \multicolumn{1}{c|}{\textbf{2.122}} 
    & \textbf{2.102} & 3.626 & \textbf{1.436} & 2.000 & \textbf{1.693} 
    & \multicolumn{1}{c|}{\textbf{2.168}} & \textbf{1.966} 
    & \textbf{3.486} & \textbf{1.457} \\
\textit{ProdLDA} & 2.000 & 1.660 & \multicolumn{1}{c|}{2.102} & 2.240 
        & \textbf{3.494} & 1.628 & 2.000 & 1.643 & \multicolumn{1}{c|}{2.099} 
        & 2.195 & 3.537 & 1.663 \\
\textit{CombinedTM}  & 2.000 & 1.622 & \multicolumn{1}{c|}{2.084} & 2.324 
            & 3.644 & 1.792 & 2.000 & 1.660 & \multicolumn{1}{c|}{2.054} 
            & 2.278 & 3.595 & 1.694 \\
\textit{BERTopic}    & \textbf{2.005} & 1.482 & \multicolumn{1}{c|}{1.9984}
            & 2.577 & 3.639 & 1.913 & \textbf{2.004} & 1.498 
            & \multicolumn{1}{c|}{2.011} & 2.595 & 3.540 & 1.836 \\ \hline
\end{tabular}%
}
\end{table}
\paragraph{Diversity} Table~\ref{tab:diversity} compares the automated diversity metric $D_{\text{unique}}$ with the LLM-based diversity rating $D_{\text{rate}}$ for both the 20NG and AGRIS datasets. For 20NG, \textit{ProdLDA} consistently shows the lowest diversity across both metrics, while \textit{LDA} exhibits relatively high diversity. \textit{BERTopic} and \textit{CombinedTM} yield intermediate scores. In AGRIS, however, the results are less consistent: for $K=50$, Mistral and Qwen rate \textit{LDA} as highly diverse, whereas Llama assigns it a lower diversity, and for $K=100$, Mistral again favors \textit{LDA} while Llama and Qwen indicate lower diversity for \textit{LDA}. Furthermore, although LLM-based evaluations generally rate \textit{BERTopic} as highly diverse, its $D_{\text{unique}}$ scores remain moderate—suggesting that even when topics share common words, contextual nuances preserve thematic distinctiveness. Overall, the divergence between $D_{\text{unique}}$ and $D_{\text{rate}}$ on AGRIS underscores the importance of considering both lexical uniqueness and semantic context when assessing topic diversity.
\begin{table}[]
    \caption{Evaluation results of diversity for 20NG \& AGRIS: automated metric $D_{\text{unique}}$ vs. LLM-based metric $D_{\text{rate}}$.}
    \label{tab:diversity}
\centering
\resizebox{\textwidth}{!}{%
\begin{tabular}{llccllccl|lccclccc}
\hline
Diversity & \multicolumn{8}{c|}{20NG} & \multicolumn{8}{c}{AGRIS} \\ \cline{2-17} 
            & \multicolumn{4}{c|}{$K=50$} & \multicolumn{4}{c|}{$K=100$} 
            & \multicolumn{4}{c|}{$K=50$} & \multicolumn{4}{c}{$K=100$} \\ \cline{2-17} 
            & \multicolumn{1}{c|}{$D_{\text{unique}}$ ↑} & \multicolumn{3}{c|}{$D_{\text{rate}}$ ↑} 
            & \multicolumn{1}{c|}{$D_{\text{unique}}$ ↑} & \multicolumn{3}{c|}{$D_{\text{rate}}$ ↑} 
            & \multicolumn{1}{c|}{$D_{\text{unique}}$ ↑} & \multicolumn{3}{c|}{$D_{\text{rate}}$ ↑}
            & \multicolumn{1}{c|}{$D_{\text{unique}}$ ↑} & \multicolumn{3}{c}{$D_{\text{rate}}$ ↑} \\ \cline{3-5} \cline{7-9} \cline{11-13} \cline{15-17} 
            & \multicolumn{1}{l|}{} & Mistral & Llama& \multicolumn{1}{c|}{Qwen} & \multicolumn{1}{l|}{} & Mistral        & Llama& \multicolumn{1}{c|}{Qwen}  & \multicolumn{1}{l|}{} & Mistral & Llama& \multicolumn{1}{c|}{Qwen} & \multicolumn{1}{l|}{} & Mistral & Llama& Qwen \\ \hline
LDA         & \multicolumn{1}{c|}{\textbf{0.636}} & \textbf{1.701} & 2.907 & \multicolumn{1}{c|}{2.884} & \multicolumn{1}{c|}{\textbf{0.614}} & \textbf{1.676} & 2.915 & \textbf{2.917}   & \multicolumn{1}{c|}{0.822} & \textbf{1.882} & 2.764 & \multicolumn{1}{c|}{\textbf{2.917}} & \multicolumn{1}{c|}{\textbf{0.795}} & \textbf{1.807} & 2.725 & 2.405 \\
ProdLDA     & \multicolumn{1}{c|}{0.307} & 1.450 & 2.793 & \multicolumn{1}{c|}{2.741} & \multicolumn{1}{c|}{0.218} & 1.417 & 2.792 & 2.731        & \multicolumn{1}{c|}{0.504} & 1.566 & 2.887 & \multicolumn{1}{c|}{2.731} & \multicolumn{1}{c|}{0.293} & 1.509 & 2.872 & 2.826 \\
CombinedTM  & \multicolumn{1}{c|}{0.454} & 1.568 & 2.912 & \multicolumn{1}{c|}{2.882} & \multicolumn{1}{c|}{0.314} & 1.555 & 2.909 & 2.871        & \multicolumn{1}{c|}{\textbf{0.871}} & 1.619 & \textbf{2.904} & \multicolumn{1}{c|}{2.871} & \multicolumn{1}{c|}{0.395} & 1.755 & 2.867 & \textbf{2.837} \\
BERTopic    & \multicolumn{1}{c|}{0.597} & 1.581 & \textbf{2.940} & \multicolumn{1}{c|}{\textbf{2.904}} & \multicolumn{1}{c|}{0.366} & 1.580 & \textbf{2.936} & \multicolumn{1}{c|}{2.903} & \multicolumn{1}{c|}{0.532} & 1.804 & 2.825 & \multicolumn{1}{c|}{2.903} & \multicolumn{1}{c|}{0.503} & 1.739 & \textbf{2.878} & 2.808 \\ \hline
\end{tabular}%
}
\end{table}
\paragraph{Irrelevant Topic Words}
Table~\ref{tab:irrelevant_topic_words} presents the evaluation of the irrelevant topic words detection metric $A_{\text{ir-topic}}$, where lower counts indicate better topic-document alignment. On 20NG, \textit{LDA}, \textit{ProdLDA}, and \textit{CombinedTM} consistently yield lower counts compared to \textit{BERTopic}, indicating that their topic words are more closely aligned with document content. 
In AGRIS, at $K=50$ two of the three LLM evaluators favor \textit{LDA} for having the fewest irrelevant words, whereas at $K=100$, \textit{CombinedTM} achieves the lowest count, suggesting its superior ability to capture nuanced document themes—likely due to its effective integration of contextual embeddings.
\begin{table}[]
    \caption{Evaluation of topic-document alignment for 20NG \& AGRIS (irrelevant topic words): LLM‐based metric $A_{\text{ir-topic}}$.} 
    \label{tab:irrelevant_topic_words}
\centering
\resizebox{\textwidth}{!}{%
\begin{tabular}{lcccccc|cccccc}
\hline
Irrelevant  & \multicolumn{6}{c|}{20NG} & \multicolumn{6}{c}{AGRIS} \\ \cline{2-13} 
Topic Words ↓ & \multicolumn{3}{c|}{$K=50$} & \multicolumn{3}{c|}{$K=100$} & \multicolumn{3}{c|}{$K=50$} & \multicolumn{3}{c}{$K=100$} \\ \cline{2-13} 
 & Mistral & Llama & \multicolumn{1}{c|}{Qwen}  & Mistral & Llama & Qwen  & Mistral & Llama & \multicolumn{1}{c|}{Qwen}  & Mistral & Llama & Qwen  \\ \hline
LDA  & \textbf{4.125} & 5.844 & \multicolumn{1}{c|}{7.676} & 4.460 & 6.011 & 7.526 & 5.287 & \textbf{5.535} & \multicolumn{1}{c|}{\textbf{8.164}} & 5.441 & 5.821 & 8.022 \\
ProdLDA     & 4.746 & \textbf{5.282} & \multicolumn{1}{c|}{\textbf{7.627}} & \textbf{3.996} & \textbf{4.605} & \textbf{6.440} & 5.121 & 5.602 & \multicolumn{1}{c|}{8.202} & 5.118 & 5.583 & 8.242 \\
CombinedTM  & 4.164 & 5.668 & \multicolumn{1}{c|}{7.895} & 4.152 & 5.922 & 8.144 & 5.132 & 5.693 & \multicolumn{1}{c|}{8.183} & \textbf{4.597} & \textbf{5.286} & \textbf{7.791} \\
BERTopic    & 4.895 & 7.362 & \multicolumn{1}{c|}{9.840} & 5.187 & 7.433 & 9.910 & \textbf{5.108} & 6.560 & \multicolumn{1}{c|}{8.578} & 5.244 & 6.748 & 8.680 \\ \hline
\end{tabular}%
}
\end{table}
\paragraph{Missing Themes} 
Table~\ref{tab:missing_themes} reports the missing themes detection metric $A_{\text{missing-theme}}$ , which quantifies the number of key document themes are omitted from the topic word list, with lower counts indicating better thematic coverage. For 20NG at $K=50$, two out of three LLM evaluators rate \textit{BERTopic} as having the lowest missing theme counts, suggesting that its topic words more comprehensively represent the document themes. At $K=100$, Qwen continues to favor \textit{BERTopic}, while both Mistral and Llama indicate that \textit{LDA} provides the best coverage. In AGRIS, however, the differences across topic models are minimal.
Overall, $A_{\text{missing-theme}}$ provides valuable insight into the extent to which topic models may fail to capture less frequent or nuanced themes from documents, which is vital for understanding long-tail phenomena and enhancing downstream applications.
\begin{table}[]
    \caption{Evaluation of topic-document alignment for 20NG \& AGRIS (missing themes): LLM-based metric $A_{\text{missing-theme}}$.}
    \label{tab:missing_themes} 
\centering
\resizebox{\textwidth}{!}{%
\begin{tabular}{lcccccc|cccccc}
\hline
Missing & \multicolumn{6}{c|}{20NG} & \multicolumn{6}{c}{AGRIS} \\ \cline{2-13} 
Themes ↓ & \multicolumn{3}{c|}{$K=50$} & \multicolumn{3}{c|}{$K=100$} & \multicolumn{3}{c|}{$K=50$} & \multicolumn{3}{c}{$K=100$} \\ \cline{2-13} 
           & Mistral & Llama& \multicolumn{1}{c|}{Qwen}  & Mistral   & Llama  & Qwen   & Mistral & Llama& \multicolumn{1}{c|}{Qwen}  & Mistral   & Llama & Qwen   \\ \hline
LDA        & \textbf{6.962}   & 7.475 & \multicolumn{1}{c|}{7.481} & \textbf{6.967} & \textbf{5.128}   & 7.694  & \textbf{4.309}   & \textbf{5.128} & \multicolumn{1}{c|}{\textbf{2.126}} & 4.458 & \textbf{5.167}  & \textbf{2.297}  \\
ProdLDA    & 7.866   & 7.739 & \multicolumn{1}{c|}{7.845} & 7.682 & 5.156   & 7.664  & 4.419   & 5.156 & \multicolumn{1}{c|}{2.676} & 4.639 & 5.438  & 2.669  \\
CombinedTM & 7.398   & 7.634 & \multicolumn{1}{c|}{6.982} & 7.347 & 5.221   & 6.974  & 4.492   & 5.221 & \multicolumn{1}{c|}{2.509} & \textbf{4.285} & 5.230  & 2.666  \\
BERTopic   & 7.582   & \textbf{7.117} & \multicolumn{1}{c|}{\textbf{5.114}} & 7.641 & 7.213   & \textbf{4.578}  & 4.662   & 5.369 & \multicolumn{1}{c|}{2.814} & 4.747 & 5.466  & 2.729  \\ \hline
\end{tabular}%
}
\end{table}
\paragraph{Divergent LLM Evaluation Patterns}
The evaluation results reveal distinct evaluation tendencies among the three LLMs, offering valuable insights for researchers using LLMs as evaluators. In terms of coherence, Qwen consistently flags a higher number of outliers $C_{\text{outlier}}$, suggesting a stricter criterion for semantic consistency, while Mistral reports lower outlier counts, indicative of a more lenient evaluation; Llama’s results generally fall between these extremes. For repetitiveness, Qwen detects fewer duplicate concepts $R_{\text{duplicate}}$ compared to Llama, with Mistral’s assessments again falling in between—demonstrating variable sensitivity to lexical redundancy across evaluators. In topic–document alignment, Qwen registers higher counts of irrelevant topic words $A_{\text{ir-topic}}$ yet lower counts of missing themes $A_{\text{missing-theme}}$ than Mistral and Llama. Moreover, all evaluators report higher missing theme counts for 20NG than for AGRIS, implying that 20NG documents exhibit greater thematic diversity and complexity. These insights underscore the influence of evaluator-specific biases on metric outcomes and highlight the importance of carefully selecting an LLM evaluator based on the intended application.
\subsection{Visualization} % revised it to 2 bigger paragarph
\begin{figure}
  \centering
  \includegraphics[width=\linewidth]{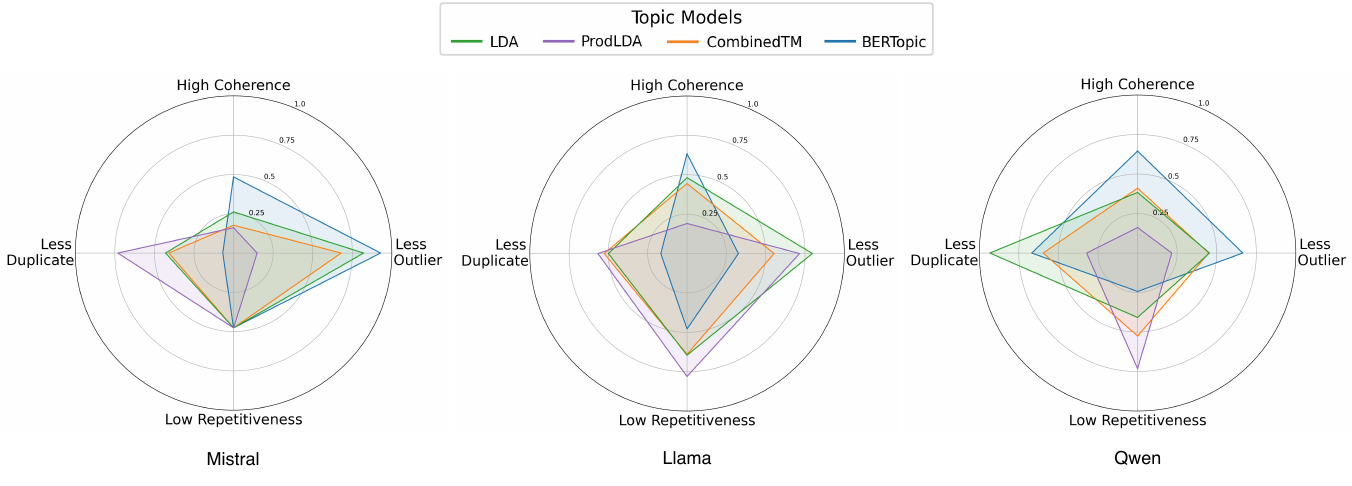}
  \caption{Radar plots comparing the evaluation trends of three LLMs in the results ($K=50$) of the 20NG}
  \label{fig:radar_20ng_50}
\end{figure}
\begin{figure}
  \centering
  \includegraphics[width=\linewidth]{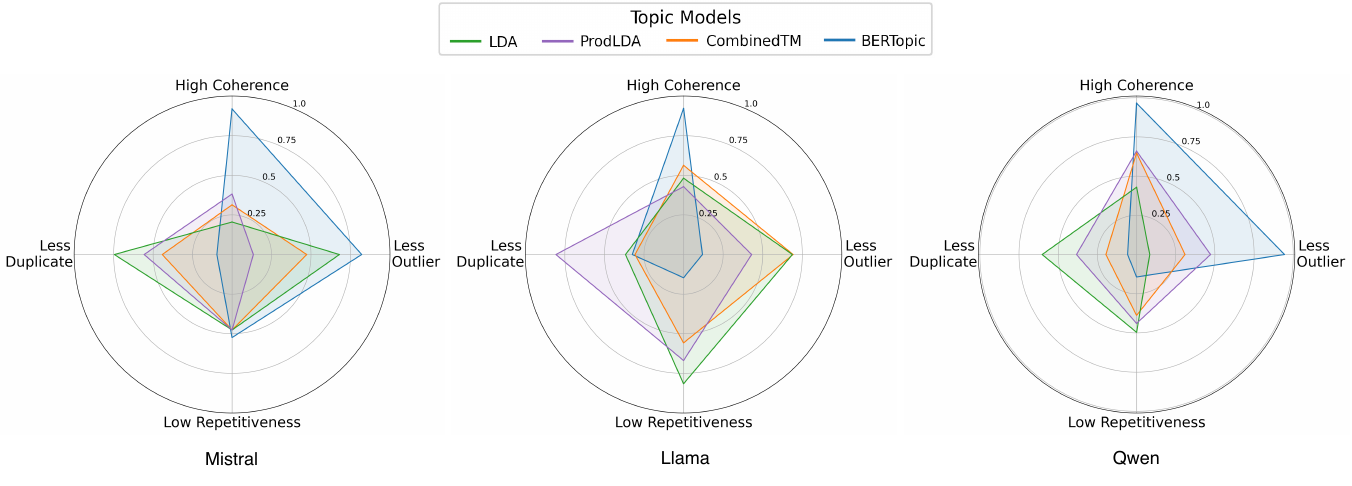}
  \caption{Radar plot comparing the evaluation trends of three LLMs in the results ($K=50$) of the AGRIS}
  \label{fig:radar_agris_50}
\end{figure}
\paragraph{Standardization of Metrics}
To enable fair comparisons across LLM evaluators (Mistral, Llama, Qwen), all metrics are normalized to the [0, 1] range using the following piecewise function:
\[
X_{\mathrm{norm}} = 
\begin{cases} 
0.5 + \dfrac{X - X_{\mathrm{mean}}}{X_{\mathrm{max}} - X_{\mathrm{min}}}, & \text{if higher values indicate better performance,} \\[1ex]
1 - \left(0.5 + \dfrac{X - X_{\mathrm{mean}}}{X_{\mathrm{max}} - X_{\mathrm{min}}}\right), & \text{if higher values indicate poorer performance.}
\end{cases}
\]
Here, \(X_{\mathrm{norm}}\) is the normalized score, while \(X_{\mathrm{mean}}\), \(X_{\mathrm{max}}\), and \(X_{\mathrm{min}}\) are the mean, maximum, and minimum values of \(X\) within each evaluator group.
\paragraph{Visualization and Analysis}
The radar plots (Figures~\ref{fig:radar_20ng_50} and~\ref{fig:radar_agris_50}) show clear discrepancies in Llama's coherence scores. A high coherence rate $C_{\text{rate}}$ should imply fewer outliers $C_{\text{outlier}}$, yet Llama rates \textit{BERTopic} high while flagging more outliers than other models. Similarly, Mistral gives low $C_{\text{rate}}$ scores to \textit{CombinedTM} and \textit{LDA} but paradoxically finds fewer outliers in their topics.

\subsection{Qualitative Analysis\label{subsec:qualitative_analysis}}
In this section, we provide a qualitative analysis of representative examples to explore discrepancies and patterns in outlier detection, duplicate concept detection.

\paragraph{Outlier Detection Discrepancies}

Outlier detection is a crucial aspect of evaluating topic coherence, as it identifies semantically inconsistent words in a topic's word list. Across the examples, outliers identified by the models often intersect but also reflect unique insights. 
Table \ref{table:example_outlier_detection} shows the examples of outlier detection discrepancies across different LLMs. 
Compared to the other two models, Mistral is more cautious in detecting outliers in topic words. On the contrary, Qwen is relatively more aggressive in detecting words with unclear semantic pointing from topic words and considering them as outliers.

\begin{table}[]
\caption{Examples of outlier detection metric $C_{\text{outlier}}$ from 20NG}
\label{table:example_outlier_detection}
\centering
\resizebox{\textwidth}{!}{%
\begin{tabular}{l|lll}
\hline
Topic words & \multicolumn{1}{l}{Mistral} & \multicolumn{1}{l}{Llama} & \multicolumn{1}{l}{Qwen} \\ \hline
interested book advance fax printer print email address mail mailing & fax & fax, printer, print & advance, fax \\
keyboard window output problem work time run input response drug & drug & drug & drug \\
science evidence theory scientific observation scientist fact explain bug claim & bug & bug & bug, claim \\ \hline
\end{tabular}%
}
\end{table}

\paragraph{Duplicate Concept Detection Contradictions}
The extracted duplicate pairs often differ significantly among the LLMs, showcasing varying thresholds for identifying conceptual overlap.
Table \ref{table:example_repetitiveness} shows that 
Mistral treats semantically related nouns (e.g., \textit{"christian"} and \textit{"church"}, collective nouns where there is intersection (e.g., \textit{"patient"} and \textit{"adult"}), and nouns that belong to the same category (e.g. \textit{"child"} and \textit{"adult"}) as conceptually identical. It has also had hallucinations (e.g., detecting a non-existent repetition of the word \textit{"customer"} for \textit{"client"} and a non-existent repetition of the word \textit{"email"} for \textit{"mail"}).
Llama treats grammatically related words (e.g., the verb “search” and its potential object “ package”), semantically opposite words (e.g., “disease” and “health “) as conceptually identical.

\begin{table}[]
\caption{Examples of duplicate concept detection metric $R_{\text{duplicate}}$ from 20NG}
\label{table:example_repetitiveness}
\centering
\resizebox{\textwidth}{!}{%
\begin{tabular}{l|lll}
\hline
Topic words & Mistral & Llama & Qwen                                   \\ \hline
\begin{tabular}[c]{@{}l@{}}faith scripture religion moral good \\ point christian church belief doctrine\end{tabular} & (faith, belief), (christian, church)    & \begin{tabular}[c]{@{}l@{}}(faith, religion), (christian, church), \\ (doctrine, scripture)\end{tabular}            & (faith, belief), (scripture, doctrine) \\ \hline

\begin{tabular}[c]{@{}l@{}}disease patient health medical child \\ year drug treatment adult number\end{tabular}      & (patient, adult), (child, adult)        & \begin{tabular}[c]{@{}l@{}}(disease, health), (patient, adult), \\ (child, patient), (treatment, drug)\end{tabular} & (disease, treatment), (medical, drug)  \\ \hline

\begin{tabular}[c]{@{}l@{}}client search directory package \\ software file mail fax database project\end{tabular}    & (client, customer), (mail, email) & \begin{tabular}[c]{@{}l@{}}(client, directory), (search, package), \\ (software, project)\end{tabular}              & None                                   \\ \hline
\end{tabular}%
}
\end{table}

\section{Conclusion}
In this work, we have introduced a comprehensive framework for evaluating topic models using LLM-based metrics that complements traditional automated metrics by incorporating nuanced measures of coherence, repetitiveness, diversity, and topic–document alignment. We designed specific evaluation protocols—including adversarial tests—to reveal not only the strengths and weaknesses of various topic models but also the intrinsic biases and judgment tendencies of different LLM evaluators. Our experiments on both the 20NG and AGRIS datasets demonstrate that LLMs can provide rich, context-sensitive insights into topic quality, while also highlighting evaluator-specific variations that are crucial for informed application in downstream tasks.

These findings illustrate the potential of our framework to expand the boundaries of topic model assessment by emphasizing both interpretability and practical application needs. Future work will focus on further refining these metrics, exploring additional LLM evaluators, and assessing how evaluator biases impact downstream tasks, thereby fostering more robust and actionable topic model assessments.
%%
%% The acknowledgments section is defined using the "acknowledgments" environment
%% (and NOT an unnumbered section). This ensures the proper
%% identification of the section in the article metadata, and the
%% consistent spelling of the heading.
\begin{comment}
\begin{acknowledgments}
  Thanks to the developers of ACM consolidated LaTeX styles
  \url{https://github.com/borisveytsman/acmart} and to the developers
  of Elsevier updated \LaTeX{} templates
  \url{https://www.ctan.org/tex-archive/macros/latex/contrib/els-cas-templates}.  
\end{acknowledgments}
\end{comment}
%%
%% Define the bibliography file to be used
\bibliography{sample-ceur}
\begin{comment}
    
%%
%% If your work has an appendix, this is the place to put it.
\appendix
appendix
\section{Online Resources}
The sources for the ceur-art style are available via
\begin{itemize}
\item \href{https://github.com/yamadharma/ceurart}{GitHub},
% \item \href{https://www.overleaf.com/project/5e76702c4acae70001d3bc87}{Overleaf},
\item
  \href{https://www.overleaf.com/latex/templates/template-for-submissions-to-ceur-workshop-proceedings-ceur-ws-dot-org/pkfscdkgkhcq}{Overleaf
    template}.
\end{itemize}
\end{comment}
\end{document}